\documentclass[conference]{IEEEtran}
\usepackage[utf8]{inputenc}
\usepackage[T1]{fontenc}
\usepackage{lmodern}

\usepackage{amsmath, amssymb, amsfonts}

\usepackage{graphicx}
\usepackage{booktabs}

\usepackage{hyperref}

\usepackage{microtype}
\usepackage{xcolor}

\usepackage[backend=biber,style=numeric]{biblatex}
\begin{document}
\title{Perspective on Bias in Biomedical AI: Preventing Downstream Healthcare Disparities}

\author{\IEEEauthorblockN{Michal Rosen-Zvi}
\IEEEauthorblockA{\textit{IBM Research - Israel} \\
\textit{Faculty of Medicine, The Hebrew University}\\
Israel \\
Michal.Rosen-Zvi@mail.huji.ac.il}
\and
\IEEEauthorblockN{Yoav Kan-Tor}
\IEEEauthorblockA{\textit{IBM Research - Israel} \\
Israel \\
yoavkt@gmail.com}
\and
\IEEEauthorblockN{Michael Danziger}
\IEEEauthorblockA{\textit{IBM Research - Israel} \\
Israel \\
Michael.Danziger@ibm.com}
\and
\IEEEauthorblockN{Agata Ferretti}
\IEEEauthorblockA{\textit{IBM Research Zurich} \\
Switzerland \\
Agata.Ferretti@ibm.com}
\and
\IEEEauthorblockN{Javier Aula-Blasco}
\IEEEauthorblockA{\textit{Barcelona Supercomputing Center} \\
Spain \\
javier.aulablasco@bsc.es}
\and
\IEEEauthorblockN{J\'ulia Falc\~ao}
\IEEEauthorblockA{\textit{Barcelona Supercomputing Center} \\
Spain \\
julia.falcao@bsc.es}
\and
\IEEEauthorblockN{Ron Shamir}
\IEEEauthorblockA{\textit{Tel Aviv University} \\
Israel \\
rshamir@tauex.tau.ac.il}
\and
\IEEEauthorblockN{Mira Marcus-Kalish}
\IEEEauthorblockA{\textit{Tel Aviv University} \\
Israel \\
miram@tauex.tau.ac.il}
\and
\IEEEauthorblockN{Mordechai Muszkat}
\IEEEauthorblockA{\textit{Hadassah University Hospital} \\
Israel\\
mordechai.muszkat@mail.huji.ac.il}
}

%\author{
%  Michal Rosen-Zvi$^{1,2}$, 
%  Yoav Kan-Tor$^{1}$, 
%  Michael Danziger$^{1}$, 
%  Agata Ferretti$^{3}$, 
%  Javier Aula-Blasco$^{4}$, 
 % J\'ulia Falc\~ao$^{4}$, 
 % Ron Shamir$^{5}$, 
 % Mira Marcus-Kalish$^{5}$, 
 % Mordechai Muszkat$^{2,6}$

%  $^1$IBM Research, Haifa, Israel \\
 % $^2$Faculty of Medicine, Hebrew University, Jerusalem, Israel \\
 % $^3$IBM Research Zurich, Switzerland \\
 % $^4$Barcelona Supercomputing Center, Spain \\
 % $^5$Tel Aviv University, Israel \\
 % $^6$Hadassah University Hospital, Israel \\
 % \\
%}

%\date{}  % removes date

\maketitle

\begin{abstract}
Healthcare disparities persist across socioeconomic boundaries, often attributed to unequal access to screening, diagnostics, and therapeutics. However, this perspective highlights that critical biases can emerge much earlier, during data collection and research prioritization, long before clinical implementation, particularly in studies focused on molecular and omics data. A vast number of studies focus on collecting omics data, but the demographic information associated with these datasets is often not reported, and when it is reported, it reveals substantial biases. An automated analysis of $4514$ PubMed-indexed omics publications from 2015 to 2024, examining reporting across multiple demographic dimensions, reveals limited reporting overall; for example, only $2.7\%$ of studies report ancestry or ethnicity information and geographic origin reporting is limited to $2.5\%$. Analysis of large-scale datasets commonly used for model training, such as CellxGene and GEO, reveals substantial population bias where European-ancestry data dominates. As biomedical foundation models become central to biomedical discovery with a paradigm in which base models are pretrained on large datasets and reusing them repeatedly for many different downstream tasks, they risk perpetuating or amplifying these early-stage biases, leading to cascading inequities that regulatory interventions cannot fully reverse.  We propose a community-wide focus on three foundational principles: Provenance, Openness, and Reliability through Evaluation Transparency. Together, these principles can help make biases and limitations more visible to model developers and users, supporting more informed model development, evaluation, and deployment decisions in biomedical AI.
\end{abstract}

\section{Introduction}\label{intro}

%\vspace{1\baselineskip}

Healthcare disparities are especially visible among the aging population, with differences in lifespan and quality of life largely shaped by socioeconomic status and geographic location. Older adults in wealthier areas tend to live longer, healthier lives compared to those in lower-income or underserved regions \cite{stringhini2017socioeconomic}. A recent study examining the elderly population in the U.S. revealed biological differences underlying racial and ethnic disparities in health, showing accelerated aging among ethnic groups from lower socioeconomic backgrounds \cite{farina2023racial}. Socioeconomic status is closely linked to healthy aging. Common explanations for why greater wealth improves health outcomes in later life include reduced stress, lower trauma exposure, decreased allostatic load, and better access to timely, appropriate healthcare in higher socioeconomic groups \cite{mcmaughan2020socioeconomic}.

While considerable attention has been given to addressing healthcare disparities at the point of care, including seminal findings that widely used clinical algorithms can encode significant racial bias \cite{obermeyer2019dissecting}, less focus has been directed toward upstream biases in the research enterprise that ultimately influence which therapeutic options become available and what guidance is provided when administering them. It is well established that, due to the high cost of drug development, the industry operates within an incentive system that prioritizes conditions affecting populations with greater ability to pay  \cite{trouiller2002drug}. The primary force behind the industry is to create and deliver a continuous stream of novel, patented, and medically attractive drugs while capturing commercial value. Due to the enormous costs associated with drug development, companies often face challenges in remaining cost-effective. Only in recent years has the high-quality availability and advanced integration of human genetics and multi-omics technologies, enabling a better understanding of disease mechanisms, begun to drive a new trend toward improved returns on investment \cite{schuhmacher2023analysis}.

We focus on the biases embedded in foundation models and their potential impact as these technologies begin to shape biomedical research and, increasingly, the broader medical domain. These models are driving new paradigms in medical AI, such as the generalist medical AI paradigm \cite{moor2023foundation}, which seeks to replace task-specific models with a unified system capable of handling diverse clinical tasks across multiple data modalities, and the virtual cell paradigm \cite{bunne2024build}, which seeks to simulate cellular behavior using generative approaches. However, the foundational datasets used to train these models often exhibit significant geographic and socioeconomic biases. For instance, individuals of European ancestry account for over $78\%$ of participants in genome-wide association studies, despite representing only $16\%$ of the global population \cite{sirugo2019missing}. If such biases are not addressed at the source, they may propagate through the research pipeline, leading to inequities that downstream interventions, even if technically sound, cannot fully correct.

This perspective is structured as follows. We begin by introducing the role of biomedical foundation models in discovery. Next, we examine the datasets these models rely on and the biases they may carry. We then discuss how explainable AI (XAI) and benchmarking tools can help assess model robustness and uncover bias. Finally, we propose a practical framework to address these challenges.

\section{The Role of Biomedical Foundation Models in Biomedical Research}\label{bmfm}
 Biomedical research focuses on understanding health and disease and finding remedies to diseases. The rationale behind the drug discovery process often involves a series of critical research steps, beginning with the identification and validation of disease-relevant targets, followed by the development and optimization of therapeutic molecules. 
 
 During clinical development, the safety and efficacy of the drug candidates are rigorously tested. Biomarkers often play a crucial role in identifying responsive patient subgroups, tailoring dosing regimens based on genetic profiles, and enabling precise patient stratification—all of which increase the likelihood of successful clinical trials while minimizing the risk of adverse effects. Since the completion of the first human genome sequence over two decades ago through the Human Genome Project, the landscape of drug discovery has been fundamentally reshaped by genomics, with a $2021$ study counting $7712$ approved and experimental pharmaceuticals associated with the advances initiated by the project \cite{gates2021wealth}. Genetic insights now play a pivotal role not only in identifying new drug targets, but also in predicting drug efficacy and safety across diverse populations. The growing recognition that genetic variation significantly influences medication response—both in terms of effectiveness and risk of adverse events—has fundamentally transformed pharmaceutical development strategies. The drug Warfarin provides a compelling illustration of this principle. As the most widely prescribed oral anticoagulant globally, Warfarin requires precise dosing due to its narrow therapeutic range and highly variable patient requirements. Although algorithms incorporating genetic information have been developed to optimize dosing, they are predominantly based on European populations. For example, genetic variants that effectively explain Warfarin dose variability in Europeans account for substantially less variability in patients of African descent. As a result, dosing algorithms derived from European datasets often fail to provide safer and more effective treatment across diverse ethnic groups \cite{asiimwe2022ethnic}. Especially for complex diseases such as Alzheimer's, it has been shown that some treatments may affect certain ancestry groups more than others~\cite{zhang-alzheimers2022}. This underscores the urgent need to identify genetic variants influencing drug metabolism across global populations to support the development of universally effective therapies.

Recent advances in omic technologies—particularly the emergence of single-cell RNA sequencing (scRNA-seq)—are adding powerful new layers of resolution to biomedical research. These technologies are reshaping approaches ranging from disease mechanism discovery and target identification to lead optimization and biomarker discovery. By pairing cutting-edge computational tools with vast public datasets, scRNA-seq enables a deeper, more precise understanding of disease biology at the cellular level, transforming how therapeutic targets are identified and validated \cite{van2023applications}.

Gene expression levels are influenced by genetic mutations in enhancer and promoter regions, as studied in transcriptome-wide association studies (TWAS) and expression quantitative trait loci (eQTL) analyses \cite{bykova-hmg2022}. Transcriptomic data has been shown to vary by ancestry group, both in healthy and cancerous cells \cite{kachuri-naturegenetics2023,arora-trendsingenetics2023}. Integrating multiple ancestry groups in TWAS analyses can improve statistical robustness \cite{chen-naturegenetics2023}. Epigenetic effects also impact cellular activity. Unlike genetic mutations, epigenetic modifications—such as DNA methylation, chromatin remodeling, nucleosome positioning, and changes in noncoding RNA profiles—are reversible and play a fundamental role in regulating gene expression.

This wealth of biological information, whether available as free text in books and journals or captured in structured forms (including genes, their variants and disease associations, proteins, protein--protein interactions, small molecules, transcriptomic data, and more), has been leveraged by numerous research groups to develop large-scale foundation models trained on uni- and multi-modal data. Examples include large language models (LLMs) trained on extensive biological corpora, such as BioGPT~\cite{luo2022biogpt} and BioBERT~\cite{lee2020biobert}; models trained on massive protein datasets to enable structure and function prediction, such as ESM-$3$~\cite{hayes2025simulating} and AlphaFold~$3$~\cite{abramson2024accurate}; models trained on DNA sequences and genomic variation, such as DNABERT~\cite{ji2021dnabert}; models trained on human transcriptomic profiles (e.g., scBERT~\cite{yang2022scbert} and BMFM-RNA~\cite{dandala2025bmfm}) or epigenomic profiles (e.g., MethylGPT~\cite{ying2024methylgpt}); and models spanning multiple domains, including multimodal and cross-species models (see, e.g.,~\cite{shoshan2024mammal,zhang2025scientific,nguyen2024sequence}).

These foundation models are increasingly used to address downstream tasks in biomedical research, including understanding disease mechanisms, discovering biomarkers for prognosis, and supporting the drug discovery process. They are becoming valuable resources in their own right. The community continues to progress toward broader data coverage and the integration of tools to create more powerful multi-omic foundation models~\cite{cui2024scgpt,cui2025towards}, alongside streamlined environments that enable agentic workflows, where intelligent agents autonomously deploy and refine these models~\cite{zheng2025learning,liu2025advances,zou2025rise}. 

Recent work suggests that agentic AI in bioinformatics can incorporate bias-awareness mechanisms, such as fairness-aware learning and multi-agent coordination, to mitigate known data biases~\cite{zhou2025streamline}. However, the effectiveness of such approaches remains dependent on the underlying data and requires explicit design and evaluation to ensure equitable outcomes across populations. As foundation models become central to biomedical innovation, their growing influence highlights the need for careful scrutiny of both the data they rely on and how these systems are deployed, in order to mitigate bias and ensure equitable outcomes.

\section{Data Used in Foundation Models: A Potential Source for Bias}\label{data}

 At the core of the success of foundation models lies the data itself—the essential fuel powering these systems. Large-scale pretraining data is key to biomedical model performance. A comprehensive list of datasets used in such models can be found in \cite{zhang2025scientific}.

In this section, we examine various data types utilized by foundation models and their potential population biases. Here, we refer to bias as the systematic under-representation of biomedical information characterizing certain groups, such as ethnic populations, age groups, or genders. We exclude "batch effects," common in omics data, arising from differences in timing, personnel, reagents, and instrumentation. These have been extensively studied (see e.g. transcriptomic/proteomic \cite{goh2017batch} and microbiome data \cite{nearing2021identifying}). As noted in \cite{goh2017batch}, batch effects often confound hidden biological heterogeneity. They are typically removed, as they lack biological meaning. This is done via correction algorithms or model architectures. For example, in single-cell RNA-seq, read depth, a key technical variable, is explicitly modeled \cite{hao2024large}.

In contrast, subpopulation effects are potentially biologically significant and should be preserved. Ignoring factors such as sex, gender, or ethnicity, and their contributions to health and disease, can lead to suboptimal results, analytical errors, and potentially discriminatory outcomes.

Demographic information is rarely reported in omics studies. To quantify this issue, we analyzed $4514$ PubMed-indexed omics publications published between 2015 and 2024 and assessed whether abstracts reported basic characteristics of the study population, including ancestry or ethnicity, geographic origin, and sex or gender composition (see Appendix \ref{sec:appendix_demographics} for methodology). Across all five major omics domains, only $2.7\%$ of publications report ancestry or ethnicity information, $4.5\%$ report geographic origin, and just $3.1\%$ report sex or gender composition. Reporting rates vary considerably by field, with genomics showing the highest ancestry reporting at $6.0\%$, while transcriptomics ($0.6\%$) and proteomics ($1.0\%$) rarely report it. These findings, summarized in Table \ref{tab:reporting}, suggest that concerns regarding population bias extend beyond the composition of individual datasets. Across omics domains, even the information needed to assess representativeness is frequently unavailable, limiting our ability to evaluate whether findings and downstream models are likely to generalize across populations. Beyond its implications for bias assessment, this lack of demographic transparency also conflicts with broader FAIR data stewardship principles, which emphasize rich metadata and provenance as prerequisites for the discoverability, interpretation, and reuse of scientific data \cite{wilkinson2016fair}.

Several limitations should be noted. First, the analysis was based on abstracts rather than full-text articles, and some demographic information may therefore have been reported elsewhere. Nevertheless, abstracts typically emphasize information that authors consider central to the study and thus provide insight into reporting priorities. Second, to assess annotation quality, we compared automated annotations against blinded human annotations for a randomly selected subset of 40 publications. While the automated approach showed a tendency toward under-detection of demographic reporting, resulting in conservative prevalence estimates, it rarely generated false positive demographic assignments and therefore is more likely to underestimate than overestimate reporting rates. Finally, our analysis focused on demographic reporting and did not assess other important determinants of health, including environmental exposures. A growing body of work demonstrates that gene–environment interactions involving lifestyle, pollutants, and socioeconomic factors substantially influence molecular phenotypes and disease risk. Consequently, the reporting gaps identified here likely represent only one component of a broader challenge in capturing the contextual information required for robust, reliable, and generalizable biomedical AI \cite{alemu2025multi,vineis2020new}.

\begin{table*}[t]
\centering
\caption{Demographic reporting rates across omics fields based on automated text mining of $4719$ PubMed-indexed abstracts (2015--2024). Values indicate the percentage of publications in each field reporting each demographic dimension.}
\label{tab:reporting}
\small
\begin{tabular}{lrrrrr}
\toprule
\textbf{Omics Field}  & \textbf{Ancestry/ } & \textbf{Geographic} & \textbf{Age} & \textbf{Sex/} & \textbf{Socio-} \\
  & \textbf{Ethnicity} & \textbf{Origin} & \textbf{Reporting} & \textbf{Gender} & \textbf{economic} \\
\midrule
Genomics    (748)    & 6.0\% & 4.7\% & 2.1\% & 0.9\% & 4.1\% \\
Transcriptomics (944)& 0.6\% & 3.6\% & 2.8\% & 2.6\% & 0.0\% \\
Proteomics     (912) & 1.0\% & 4.8\% & 2.2\% & 1.1\% & 0.0\% \\
Epigenomics    (917) & 4.0\% & 5.0\% & 6.2\% & 4.3\% & 1.7\% \\
Microbiome     (993) & 2.3\% & 4.2\% & 7.6\% & 6.1\% & 0.9\% \\
\midrule
\textbf{All fields} & \textbf{2.7\% (120)} & \textbf{4.5\% (201)} & \textbf{4.3\% (194)} & \textbf{3.1\% (142)} & \textbf{1.2\% (56)}  \\
\bottomrule
%\\[-0.5em]
%\multicolumn{7}{l}{\footnotesize \textit{Temporal trend (ancestry reporting):} 2.0\% (2015--2019) $\rightarrow$ 2.8\% (2020--2022) $\rightarrow$ 3.7\% (2023--2024)} \\
\end{tabular}
\end{table*}

\textbf{Genomics}: As previously discussed, models designed to learn from genomic variation often rely on whole genome sequencing data obtained from individuals. However, the data available to date has been shown  to have a skewed population distribution, over-representing certain groups while under-representing others \cite{fatumo2022diversity}. This imbalance limits the model’s ability to generalize across diverse ancestries and may result in biased genetic risk predictions or missed therapeutic opportunities for underrepresented populations. For example, polygenic risk scores estimate an individual’s genetic risk for complex diseases by combining the effects of many genetic variants identified through GWAS. However, as noted above, most existing GWAS data comes from individuals of European ancestry, and studies have shown that the accuracy of risk scores derived from such data declines as the genetic distance from the GWAS population to the target population increases \cite{martin2019clinical}. This situation may deepen health disparities between ancestral groups.

\textbf{Transcriptomics}: Bulk RNA-seq, and more recently, single-cell RNA-seq (as well as single-nucleus RNA-seq), have become key research tools in biomedical research. Large databases, such as CellxGene, have emerged to support this work \cite{the2022tabula,czi2025cz}. Treatment targets are often selected based on studies that use these data to uncover the mechanisms underlying various diseases. While CellxGene includes structured metadata fields at both the cell and dataset levels, such as sex and self-reported ethnicity, these annotations are inconsistently reported and remain incomplete across a substantial fraction of datasets. Consequently, the presence of metadata does not translate into representative coverage, and CellxGene, the largest known source of human single-cell RNA-seq data to date, remains skewed toward samples from adults of European or unknown ethnicity \cite{czi2025cz}.

\textbf{Epigenomics}: Recent efforts have focused on learning from DNA methylation, ChIP-seq, and histone modification data, which offer insights into gene regulation and cellular identity \cite{camillo2024cpgpt}. These datasets, however, also suffer from population and tissue sampling biases. A related repository that collects transcriptomic and epigenomic data is the Gene Expression Omnibus (GEO) \cite{clough2016gene}. A recent study assessed whether GEO datasets reflect the diversity of the U.S. population using DEI criteria (gender, ethnicity, ancestry, race). The findings revealed that while many datasets displayed balanced gender ratios, significant bias in ethnicity representation was observed, with a predominance of White and African American participants \cite{gondal2024assessing}.

\textbf{Proteomics}: Similarly, protein-focused models face their own limitations. Despite the large number of known proteins, only a small subset is actively studied as potential drug targets \cite{carter2019target}. Unlike genomic or transcriptomic data, the selection of proteins for study is often dictated by the availability of biological tools—such as antibodies or promoters—that enable functional investigation in laboratory settings, rather than by considerations of population diversity.

\textbf{Microbiome}: Microbiome profiles from various human body niches (e.g., gut, vagina, airways) and environmental sources have become widely accessible, thanks to the efforts of many research teams. These datasets span diverse age groups, genders, disease conditions, and anatomical sites. While computational methods have long been applied to microbiome data, only recently have foundation models demonstrated the ability to enable precise and efficient microbiome-based classification and prediction \cite{han2025techniques}. A large-scale metagenomic assembly approach aimed at reconstructing bacterial and archaeal genomes across multiple populations, body sites, and host ages revealed substantial phylogenetic and functional diversity—much of which remains uncaptured, particularly from rare organisms, non-stool sample types, underrepresented global populations, and varied lifestyles \cite{pasolli2019extensive}. 

\textbf{Texts}: Pre-trained language models are also subject to bias. Studies have shown that these models tend to over-represent dominant cultural and scientific perspectives, while encoding biases that may marginalize underrepresented groups \cite{bender2021dangers}. In particular, a recent systematic review revealed pervasive demographic biases in LLMs trained on clinical notes, medical literature and guidelines, with gender and racial/ethnic biases being particularly common \cite{omar2025evaluating}. This concern extends to biomedical applications, where language models trained on biased literature may perpetuate an overemphasis on data from individuals of European descent. Consequently, such models risk reinforcing existing disparities in research and healthcare by under-representing findings relevant to diverse populations. 

To summarize, it is well known that current datasets contain biases. Modern large models, with hundreds of millions or even billions of parameters trained on massive datasets, make identifying and tracking these biases particularly challenging. Their apparent success may be misleading and may be mistakenly perceived as delivering objective truth.

Because all datasets are collected under practical, scientific, and historical constraints, no dataset can fully capture the diversity of biological systems and patient populations. Consequently, transparency regarding data provenance, study populations, and model behavior becomes critically important. Making these factors more visible can help researchers better understand the scope and limitations of foundation models, assess the contexts in which they are likely to be reliable, and identify situations in which additional validation may be warranted. The next section discusses methods for uncovering and characterizing model bias.

\section{Interrogating the Models: XAI and Benchmarks}\label{benchmarks}

Explainable AI (XAI) technologies can support developers in interrogating their models and uncovering potential biases. These technologies generally fall into three broad categories \cite{malinverno2023historical}: causal inference methods, inherently explainable models, and post hoc analysis techniques.

Causal inference methods use AI to infer causal relationships and perform root-cause analysis. These approaches often assume that causal relationships are known and apply weighting techniques to estimate causal effects. A recent paper \cite{zhang2024towards} demonstrated how the attention mechanisms in foundation models can be leveraged to address causal questions at a larger scale and with greater generalizability than traditional statistical tools. This emerging direction toward causally aware foundation models is expected to gain increasing attention \cite{liu-etal-2025-large-language}.

Inherently explainable models, such as decision trees, offer transparency in their decision-making processes, making them more interpretable to human users. Post hoc analysis techniques aim to explain the outputs of complex, black-box models after training. Unfortunately, foundation models often suffer from poor interpretability due to their intricate and opaque architectures. A recent study on explainability in LLMs reviews several established methods, such as SHAP and attention mechanisms, while also pointing out their inherent limitations \cite{mumuni2025explainable}. SHAP (SHapley Additive exPlanations), a game-theoretic technique, assigns an importance value to each feature in a prediction, thereby quantifying the contribution of individual inputs to the model’s output. Furthermore, research has demonstrated that the data generation process itself can influence the resulting explanations, underscoring the critical need to examine and understand this process \cite{mhasawade2024understandingdisparitiesposthoc}.

Explainability methods may also contribute to the identification of population-related biases in biomedical foundation models. For example, if a model trained predominantly on data from a specific population systematically relies on molecular features, genetic variants, or clinical correlates that are unevenly represented across populations, feature attribution approaches such as SHAP can help reveal these dependencies. Similarly, attention-based analyses may identify biological signals that consistently drive model predictions, while causal approaches may help distinguish population-associated correlations from biologically meaningful mechanisms. Although explainability alone cannot establish the presence or absence of bias, it can provide an additional layer of transparency that, when combined with rich metadata and provenance information, helps researchers evaluate whether model behavior is likely to generalize across diverse populations.

There is a clear need for further development of post hoc methods specifically tailored to foundation models. Interestingly, LLMs themselves may be leveraged to help explain the reasoning behind the outputs of other black-box biomedical models. However, to do so effectively, these LLMs must possess sufficient domain expertise in biomedicine.

A widely used approach for evaluating algorithmic performance in data science is the common task framework. In this setup, participants develop models using publicly available training datasets to perform a shared task—such as class prediction or content generation. These models are then submitted to a scoring system (the referee), which evaluates them on a hidden test dataset \cite{donoho201750}. 

To assess biases in biomedical models more effectively, we advocate for the development of targeted common tasks and benchmarks.  A recent meta-assessment of single-cell analysis benchmarks found that, although many are reproducible and make their code publicly available, they are often difficult to extend—particularly as new methods and evaluation criteria emerge \cite{sonrel2023meta}.   Moreover, to the best of our knowledge, none of the currently available benchmarks are specifically designed to rigorously evaluate aspects related to population-based biases.  %Here we present three examples of recently developed benchmarks that either already address some related aspects or appear well positioned for such investigation.

There are already platforms specifically designed to simplify the creation and extension of benchmarks. For example, the \textbf{Open Problems in Single-Cell Analysis} platform is an open-source, extensible, and continuously evolving benchmarking framework that enables rigorous, quantitative evaluation of best practices in single-cell analysis \cite{luecken2025defining}. Another example is \textbf{BenchEMark} \cite{capella2017lessons}, a community-driven infrastructure built to support continuous, automated benchmarking of bioinformatics methods, tools, and web services. These platforms provide a strong foundation for designing new benchmarks that explicitly assess population-level disparities. Encouragingly, the community is already moving in this direction—laying the groundwork for more inclusive, transparent, and bias-aware evaluation practices in biomedical AI.

\section{Recommendations}\label{recom}

Data paucity remains a pervasive challenge in efforts to capture biological complexity through pretraining biomedical foundation models on raw laboratory data. While initiatives aimed at diversifying biomedical datasets, such as inclusive genome sequencing roadmaps \cite{fatumo2022diversity} and proteome exploration efforts \cite{carter2019target}, are essential, they are often resource-intensive and slow to implement. In the meantime, this paper advocates for complementary strategies that focus on making existing biases, and the complex ways in which they influence early stages of foundation model development, more visible, thereby raising awareness and enabling more informed  choices. 

We recommend that the foundation model development community collaborate with key stakeholders across the biomedical ecosystem, including clinicians, patients, researchers, industry leaders, and regulators, to consistently work along the following three dimensions:

\textbf{Provenance}: Capturing and reporting the origin of data, including ethnicity, geography, and socioeconomic context, is essential for assessing model generalizability and identifying population-specific risks. Without this transparency, AI tools risk reinforcing existing disparities. The CONSORT Guidelines \cite{turner2012consolidated}, widely used in clinical trials, offer a model for how provenance reporting could be standardized in biomedical AI.

\textbf{Openness}: Open science practices and collaborative initiatives are vital for democratizing access to tools, datasets, and evaluation protocols. These efforts foster transparency, reproducibility, and accountability in AI development.

\textbf{Reliability, Benchmarking and Evaluation Transparency}: Ensuring the reliability of biomedical AI systems, including their robustness, consistency across populations, and stability under distribution shifts, is critical for their safe and effective use. Current benchmarks rarely assess population-level biases. We advocate for the creation of new, inclusive benchmarks that explicitly evaluate model performance across diverse subgroups. Transparent reporting of evaluation metrics, especially those related to fairness and equity, should become standard practice.

By embedding these principles into the AI development pipeline, we can improve the visibility of biases and limitations before models are widely deployed. This proactive approach complements regulatory oversight and helps promote more transparent, reliable, and appropriately evaluated biomedical AI systems.

%\subsection{Figures}

%\begin{figure}
 % \centering
  %\fbox{\rule[-.5cm]{0cm}{4cm} \rule[-.5cm]{4cm}{0cm}}
  %\caption{Sample figure caption.}
%\end{figure}

%All artwork must be neat, clean, and legible. Lines should be dark enough for
%purposes of reproduction. The figure number and caption always appear after the
%figure. Place one line space before the figure caption and one line space after
%the figure. The figure caption should be lower case (except for first word and
%proper nouns); figures are numbered consecutively.

%\section*{References}
\section*{Acknowledgments}
The research leading to these results has received funding from the Horizon Europe Programme under the Marie Skłodowska-Curie grant agreement No 101183031. Views and opinions expressed are however those of the author(s) only and do not necessarily reflect those of the European Union. Neither the European Union nor the granting authority can be held responsible for them.

\printbibliography
\appendix

\section{Demographic Analysis Details}
\label{sec:appendix_demographics}

\subsubsection{Study Design and Data Source}
To quantify demographic reporting practices in omics research, we conducted a systematic analysis of PubMed abstracts published between January 2015 and December 2024. We queried the PubMed database using the NCBI Entrez Programming Utilities via the Biopython library (version 1.87), adhering to NCBI usage guidelines.

\subsubsection{Search Strategy}
We analyzed five major omics domains: genomics, transcriptomics, proteomics, epigenomics, and microbiome research. For each field, we searched for publications containing the respective term (e.g., ``genomics'') in the title or abstract and indexed with the ``human'' Medical Subject Heading (MeSH) term. Publications were stratified into three time periods: 2015--2019, 2020--2022, and 2023--2024. We retrieved up to 350 top-ranked abstracts (by relevance) per field per time period, yielding $4719$ abstracts with complete data ($89.9\%$ retention rate from $5250$ retrieved). Further analysis was performed to identify duplications. $205$ duplications were found leading to a starting cohort of $4514$ papers.

\subsubsection{Text Mining and Classification}
We assessed five demographic dimensions using keyword-based pattern matching with regular expressions: (1) ancestry/ethnicity (64 patterns), (2) geographic origin (9 patterns), (3) age reporting (11 patterns), (4) sex/gender (10 patterns), and (5) socioeconomic status (10 patterns). Keywords included terms such as  ``multi-ethnic,'' ``European ancestry,'' ``participants from [location],'' ``age range,'' ``sex distribution,'' and ``socioeconomic status.'' Each abstract was classified as reporting (present) or not reporting (absent) each dimension based on case-insensitive pattern matching in the combined title and abstract text. 

\subsubsection{AI-Assisted Analysis Development}
The analytical framework and scientific questions were defined by the authors. Large language models were used as programming and research assistants to accelerate implementation, including code generation, debugging, and development of candidate extraction patterns (IBM's BoB (Build on Bedrock) and Anthropic's Claude Opus 4.6). All extraction rules, keyword dictionaries, validation procedures, and final methodological choices were reviewed and approved by the authors. The resulting analysis was performed using a deterministic rule-based text-mining pipeline implemented in Python, ensuring that all reported results are fully reproducible and independent of the behavior of any specific language model.

\subsubsection{Statistical Analysis}
Reporting rates were calculated as the percentage of abstracts containing at least one keyword for each demographic dimension. We analyzed temporal trends across the three time periods and field-specific patterns across the five omics domains. Cross-tabulations examined the interaction between field and time period.

\subsubsection{Validation of the Automated Text-Mining Pipeline}
To assess the reliability of the automated extraction approach, we conducted a blinded validation study on a randomly selected subset of 40 publications from the analyzed corpus. Human expert annotations were independently generated and compared against the automated annotations for both omics field classification and demographic reporting detection.

For omics field classification, the automated pipeline demonstrated substantial agreement with human annotation Cohen’s $\kappa  = 0.78$. Only a single complete disagreement was observed. All discrepancies occurred in studies involving multiple omics modalities, reflecting the inherent ambiguity of assigning a single omics category to multimodal studies rather than systematic classification errors.

For demographic reporting detection, the automated pipeline achieved high specificity $97.8\%$  and an overall agreement of $91\%$ with human annotations. However, recall was limited ($12.5\%$), indicating that the system was substantially more likely to miss demographic information than to report it incorrectly. Examination of the validation set suggests that some false negatives occurred when demographic characteristics were implied by the study population description rather than explicitly reported.  The ratio of false negatives to false positives was approximately 3.5:1, demonstrating a systematic tendency toward under-detection rather than over-detection. Geographic origin and sex/gender reporting achieved fair agreement with human annotation (Cohen’s $\kappa = 0.29$), whereas ancestry/ethnicity, age reporting, and socioeconomic indicators showed particularly low recall in the validation set.

\subsubsection{Limitations}
This analysis has several limitations. First, keyword-based pattern matching may miss implicit demographic information or alternative terminology. Second, we analyzed abstracts only; full-text articles may contain additional demographic details. Third, our binary classification (present/absent) does not capture the quality or completeness of reporting. Fourth, we used simplified single-term queries (e.g., ``genomics'') rather than comprehensive Boolean searches, potentially affecting the representativeness of our sample. Finally, our analysis was limited to English-language publications indexed in PubMed.

\end{document}